\documentclass[runningheads]{llncs}
\usepackage[T1]{fontenc}
\usepackage{graphicx}
\usepackage{caption,subcaption}
\usepackage{multirow}
\usepackage[table]{xcolor}
\usepackage{url}
\usepackage{soul}

\setstcolor{red}

\usepackage{color}

\usepackage{graphicx}
\usepackage{booktabs}
\usepackage{pbox}
\usepackage{setspace}

\begin{document}
\title{Rehabilitation Exercise Quality Assessment and Feedback Generation Using Large Language Models with Prompt Engineering}
\titlerunning{Rehabilitation Exercise Quality Assessment and Feedback}

\author{
Jessica Tang\inst{1,2}\orcidID{0009-0009-7725-6702}\and
Ali Abedi\inst{1}\orcidID{0000-0002-7393-1362} \and
Tracey J.F. Colella\inst{1}\orcidID{0000-0001-6228-9012} \and
Shehroz S. Khan\inst{1,3}\orcidID{0000-0002-1195-4999}
}

\authorrunning{Tang et al.}
%
\institute{KITE Research Institute, Toronto Rehabilitation Institute, University Health Network, Toronto, Canada 
\and
Faculty of Applied Science and Engineering, University of Toronto, Toronto, Canada\\
\and
College of Engineering and Technology, American University of the Middle East, Egaila, Kuwait\\
\email{jessicao.tang@mail.utoronto.ca, \{ali.abedi,tracey.colella,shehroz.khan\}@uhn.ca}
}

\maketitle              
\begin{abstract}
Exercise-based rehabilitation improves quality of life and reduces morbidity, mortality, and rehospitalization, though transportation constraints and staff shortages lead to high dropout rates from rehabilitation programs. Virtual platforms enable patients to complete prescribed exercises at home, while AI algorithms analyze performance, deliver feedback, and update clinicians. Although many studies have developed machine learning and deep learning models for exercise quality assessment, few have explored the use of large language models (LLMs) for feedback and are limited by the lack of rehabilitation datasets containing textual feedback. In this paper, we propose a new method in which exercise-specific features are extracted from the skeletal joints of patients performing rehabilitation exercises and fed into pre-trained LLMs. Using a range of prompting techniques, such as zero-shot, few-shot, chain-of-thought, and role-play prompting, LLMs are leveraged to evaluate exercise quality and provide feedback in natural language to help patients improve their movements. The method was evaluated through extensive experiments on two publicly available rehabilitation exercise assessment datasets (UI-PRMD and REHAB24-6) and showed promising results in exercise assessment, reasoning, and feedback generation. This approach can be integrated into virtual rehabilitation platforms to help patients perform exercises correctly, support recovery, and improve health outcomes.

\keywords{Rehabilitation Exercise, Action Quality Assessment, Large Language Models, Feedback Generation, Prompt Engineering.}
\end{abstract}

\section{Introduction}
\label{sec:introduction}
Patients recovering from a cardiac event, stroke, or other traumatic injuries are often referred to rehabilitation programs to support faster recovery. These programs aim to enhance patients' quality of life by promoting independent living and reducing the likelihood of hospital readmissions, morbidity, and mortality \cite{WHORehabilitation}. Typically, rehabilitation programs focus on exercises designed to restore mobility, rebuild muscle mass, and improve overall strength \cite{dibben2023exercise}. Traditionally delivered in clinical settings or institutional environments, these programs frequently face challenges such as long wait times, limited staffing, and logistical barriers to participation, including transportation difficulties and scheduling constraints \cite{shirozhan2022barriers}. Virtual and home-based rehabilitation programs \cite{ferreira2023usage} offer a practical alternative, addressing these challenges while providing benefits comparable to in-person care \cite{seron2021effectiveness,boukhennoufa2022wearable}. By analyzing data collected during virtual sessions, Artificial Intelligence (AI) can be used to assess exercise quality, monitor patient progress, and predict program dropout risks \cite{abedi2024artificial}. These AI-driven methods typically leverage various sensors, such as wearable devices and cameras, to track patient movements. AI algorithms process this information in real time \cite{ferreira2023usage,sardari2023artificial}, offering valuable insights into exercise performance and enabling healthcare professionals to effectively monitor and personalize patient care interventions \cite{sardari2023artificial,abedi2024artificial}.

AI-driven methods for assessing the quality of rehabilitation exercises primarily rely on three types of data: acceleration data from inertial wearable sensors, video data from RGB or depth cameras, and body joint data \cite{sardari2023artificial,ettefagh2024technological,brennan2019feedback}. Body joint data, in particular, is either captured using sensors such as Kinect or extracted from RGB videos using computer vision techniques \cite{sardari2023artificial,pavllo20193d,lugaresi2019mediapipe,abedi2023cross}. Prior research in general human activity analysis has emphasized the importance of body joint analysis in enhancing performance recognition \cite{yan2018spatial}. In the context of rehabilitation, body joint analysis closely aligns with clinical practices used to evaluate exercise technique and quality \cite{kimore}. Moreover, compared to video data, body joint data has lower dimensionality and is less affected by variations in lighting and background, making it a more robust and reliable modality for analysis. This paper focuses on evaluating rehabilitation exercises through the analysis of body joint sequences.

Real-time feedback on rehabilitation exercise quality and technique \cite{ettefagh2024technological,brennan2019feedback} plays a crucial role in enhancing the effectiveness of exercise-based rehabilitation. It not only helps correct movement execution and ensure proper technique but also motivates patients and strengthens their psychological resilience \cite{schuber2024relevance,popovic2014feedback,abedi2024artificial}. While many studies have explored body joint–based assessment of exercise quality \cite{sardari2023artificial} and feedback generation \cite{ettefagh2024technological,brennan2019feedback}, the potential of Large Language Models (LLMs) to deliver rich, personalized, and context-aware feedback remains largely underexplored \cite{wang2024ubiphysio}.

This paper investigates the applicability of Large Language Models (LLMs) for assessing rehabilitation exercise quality and generating feedback, making the following key contributions: (1) it proposes a novel framework that enables pre-trained LLMs to evaluate rehabilitation exercise quality by integrating exercise-specific features with prompt engineering techniques; and (2) it conducts extensive experiments on two publicly available rehabilitation exercise datasets, the University of Idaho-Physical Rehabilitation Movements Dataset (UI-PRMD) \cite{prmd} and the REHAB24-6 dataset \cite{vcernek2024rehab24}, demonstrating the feasibility of using pre-trained LLMs for both exercise quality assessment and natural language feedback generation.


\section{Related Work}
\label{sec:related_work}

This section briefly reviews existing approaches for rehabilitation exercise quality assessment \cite{sardari2023artificial} and feedback generation on exercise quality \cite{ettefagh2024technological,brennan2019feedback}.

\subsection{Rehabilitation Exercise Quality Assessment}
\label{sec:related_work_assessment}

Liao et al. \cite{liao2020deep} assessed rehabilitation exercise quality using principal component analysis for dimensionality reduction and Long Short-Term Memory (LSTM) autoencoders. Their model employed temporal pyramid sub-networks with 1D convolutions on joint sequences at different time resolutions, followed by LSTM layers for quality evaluation.

Abedi et al. \cite{abedi2023cross} proposed a rehabilitation exercise quality assessment method using MediaPipe for joint extraction, followed by LSTM models trained on exercise-specific features \cite{guo2021exercise}. To enhance generalizability, they applied cross-modal video-to-body-joint augmentation. Karagoz et al. \cite{karagoz2023supervised} further improved this approach with supervised contrastive learning to address imbalanced exercise sample distributions.

Deb et al. \cite{deb2022graph} applied Spatial-Temporal Graph Convolutional Networks (ST-GCNs) for rehabilitation exercise quality assessment, replacing global average pooling with an LSTM layer to enhance performance. Zheng et al. \cite{zheng2023skeleton} improved the robustness of ST-GCNs by introducing a rotation-invariant descriptor. Réby et al. \cite{reby2023graph} combined ST-GCNs with transformers by incorporating spatial and temporal self-attention modules, though their approach did not outperform the baseline ST-GCN model \cite{yan2018spatial}. Karlov et al. \cite{karlov2024rehabilitation} further advanced ST-GCNs by integrating contrastive learning with hard and soft negatives to develop an exercise-agnostic model for more effective assessment. Most recently, Bruce et al. \cite{bruce2024egcn++} proposed an Ensemble-based Graph Convolutional Network (EGCN++), which integrates position and orientation features through a novel fusion strategy, achieving improved performance in rehabilitation exercise quality assessment.

Apart from deep learning-based methods, threshold-based approaches have also been used for rehabilitation exercise quality assessment; however, they generally underperform compared to deep learning models due to their limited ability to handle variability in human movement and sensor noise \cite{ettefagh2024technological}. While these methods can evaluate exercise quality through classification or regression (i.e., by assigning a class label or quality score), they do not generate natural language feedback to guide patients in correcting their exercise performance.

\subsection{Rehabilitation Exercise Feedback Generation}
\label{sec:related_work_feedback}
Brennan et al. \cite{brennan2019feedback} and Ettefagh and Roshan Fekr \cite{ettefagh2024technological} reviewed AI-driven rehabilitation exercise feedback and found most studies used inertial sensor data, with some leveraging Kinect-captured body joint data. Feedback was primarily delivered via animated avatars through audio, visual, haptic, or multimodal channels. Feedback was categorized into knowledge of results (e.g., repetition count) and knowledge of performance, which can be descriptive (error explanation) or prescriptive (correction guidance) \cite{parker2011review}. However, neither review identified the use of LLMs for generating rehabilitation exercise feedback.

Wang et al. \cite{wang2024ubiphysio} introduced the first LLM-enabled platform for rehabilitation exercise feedback generation. They collected data from 104 subjects using inertial sensors, annotated with physiotherapist-guided corrections. Features extracted from the data, combined with an action token and a prompt, were used to fine-tune LLMs for feedback generation. Among the reviewed works, Wang et al. \cite{wang2024ubiphysio} is most closely related to the present study, as it represents an initial attempt to apply LLMs for rehabilitation exercise feedback. However, their approach required resource-intensive fine-tuning due to the lack of advanced prompting strategies, limiting its scalability and practical deployment. In contrast, the method proposed in this paper addresses this gap by demonstrating that strategically designed prompting can effectively enable LLMs to generate rehabilitation feedback without the need for costly model fine-tuning, thereby offering a more efficient and accessible solution.

\section{Method}
\label{sec:method}
\begin{figure}[h]
\centering
\includegraphics[width=.9\textwidth]{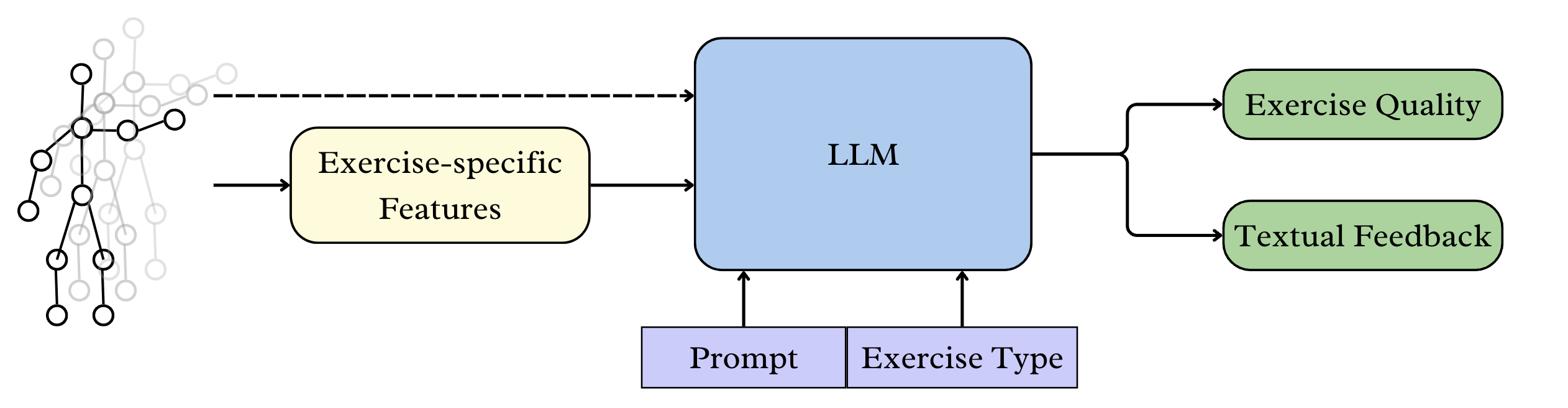}
\caption{Either body joint data or exercise-specific features extracted from body joints, combined with engineered prompts and exercise type, are fed into a pre-trained LLM. The LLM then generates exercise quality assessments and provides textual feedback.}
\label{fig:pipeline_diagram}
\end{figure}

Figure \ref{fig:pipeline_diagram} illustrates the proposed method. The input consists of a body joint sequence representing a subject performing one repetition of a rehabilitation exercise. The output includes an assessment of exercise quality and textual feedback on the performed movements. If the movements are incorrect, the feedback provides guidance on how to correct them to ensure proper execution of the exercise. Either the raw body joint sequence or a set of exercise-specific features extracted from the joints, along with a prompt and exercise type, is fed into a pre-trained LLM to generate both the quality assessment and the corresponding feedback.

The dimensionality of body joint sequences varies depending on the acquisition device. For example, in the UI-PRMD dataset \cite{prmd}, body joint data were extracted using a Kinect depth camera, capturing 22 joints with three spatial coordinates ($x$, $y$, $z$). Similarly, the REHAB24-6 dataset \cite{vcernek2024rehab24}, which employed multiple wearable inertial measurement units, contains data for 26 joints, each with three channels ($x$, $y$, $z$). Each data sample represents a single repetition of an exercise and has a dimensionality of $num\_frames \times num\_joints \times num\_channels$.

\subsection{Feature Extraction}
\label{sec:feature_extraction}

The rehabilitation clinicians involved in designing the rehabilitation exercise program provided guidelines on how the exercises should be performed correctly and established criteria for identifying incorrect movements \cite{prmd,vcernek2024rehab24,li2024finerehab,kimore,irds}. These guidelines and criteria were incorporated into the proposed method to extract exercise-specific features from body joint sequences. For instance, in the side lunge rehabilitation exercise in UI-PRMD, the correct movement is defined as "the subject takes a step to the side and lowers the body toward the floor" \cite{prmd}. The non-optimal movements classified as incorrect include: "moderate to significant knee valgus collapse, pelvis dropping or rising more than 5°, trunk angle less than 30°, thigh angle exceeding 45°, and the center of the knee positioned anterior to the toes" \cite{prmd}. Based on these criteria, the proposed method extracted three features from body joints at each frame: knee valgus angle, thigh angle, and pelvic stability. This resulted in a \(num\_frames \times num\_features \) array, where $num\_features$ is 3. Information about the patients' dominant side was incorporated into the feature extractor for movements that are performed one side at a time. Tables \ref{tab:features} (a) and (b) outline the rehabilitation exercises included in UI-PRMD and REHAB24-6, respectively, along with three to five exercise-specific features extracted for each exercise as part of the proposed method. The GitHub repository\footnote{https://github.com/jessicaxtang/exercisellm} released with this paper contains code for feature extraction from these datasets.

\begin{table}[h]
\centering
\caption{Rehabilitation exercises from (a) UI-PRMD (m01–m10) and (b) REHAB24-6 (ex1–ex6) datasets with exercise-specific feature descriptions ("A." denotes Angle).}
\label{tab:features}
\resizebox{\textwidth}{!}{%
\begin{tabular}{llp{0.7\textwidth}}
\toprule
\multicolumn{3}{l}{\textbf{(a) UI-PRMD}} \\ 
\midrule
\textbf{\#} & \textbf{Exercise} & \textbf{Extracted Features} \\ 
\midrule
m01 & Deep squat & Knee Flexion A., Hip Flexion A., Trunk Inclination A. \\ 
m02 & Hurdle step & Trunk Inclination A., Hip Flexion A., Leg Height \\ 
m03 & Inline lunge & Front Knee A., Back Knee A., Trunk Inclination A., Foot Distance \\ 
m04 & Side lunge & Knee Valgus A., Thigh A., Pelvic Stability \\ 
m05 & Sit to stand & Trunk Inclination A., Hip Flexion A., Pelvic Stability \\ 
m06 & Active straight leg raise & Hip Flexion A., Leg Elevation A., Pelvic Stability \\ 
m07 & Shoulder abduction & Arm Elevation A., Elbow Flexion A., Torso Inclination A. \\ 
m08 & Shoulder extension & Shoulder Extension A., Head Neutral Position, Trunk Inclination A. \\ 
m09 & Shoulder internal-external rotation & Arm Internal Rotation A., Arm External Rotation A., Elbow Flexion A. \\ 
m10 & Shoulder scaption & Arm Elevation A., Trunk Inclination A., Arm Plane Deviation \\ 
\midrule
\multicolumn{3}{l}{\textbf{(b) REHAB24-6}} \\ 
\midrule
\textbf{\#} & \textbf{Movement} & \textbf{Extracted Features} \\ 
\midrule
ex1 & Arm Abduction & Arm Elevation A., Trunk Inclination A., Elbow A., Plane Deviation \\ 
ex2 & Arm VW & V-Shape A. (shoulder), W-Shape A. (elbow), Trunk to Vertical A. \\ 
ex3 & (Inclined) Push-ups & Elbow Flexion A., Trunk Inclination A., Hand Symmetry, Pelvic Stability \\ 
ex4 & Leg Abduction & Leg Elevation A., Trunk A., Pelvic Tilt A., Knee A., Leg Plane Deviation \\ 
ex5 & Leg Lunge & Front Knee A., Back Knee A., Trunk A., Foot Distance \\ 
ex6 & Squats & Knee Flexion A., Hip Flexion A., Trunk A., Foot Symmetry \\ 
\bottomrule
\end{tabular}%
}
\end{table}

\subsection{Prompt Engineering}
\label{sec:prompt}
A variety of prompting techniques \cite{zaghir2024prompt} guide the LLM in analyzing feature sequences derived from rehabilitation exercise body joint data, assessing exercise quality, and generating textual feedback. The initial prompt, aimed at \textit{classifying} exercises as correct and incorrect exercises, served as the basis for more advanced prompts that were subsequently developed (Table \ref{tab:prompt}).

The classification prompt was initially evaluated across different shot settings, ranging from \textit{zero-shot} to \textit{few-shot} prompting \cite{brown2020language}. Also known as $k$-shot prompting, this approach refers to the number of labeled examples ($k$) provided per class in classification tasks. In \textit{zero-shot} prompting ($k=0$), the LLM is presented with only a task description or question, relying solely on its pre-trained knowledge to generate a response. In contrast, \textit{few-shot} prompting ($k \in \{1, 2, 3, 4\}$) improves performance by incorporating a small set of labeled examples into the input prompt, allowing the model to infer the task's structure and apply it effectively \cite{zaghir2024prompt,touvron2023llama}. The optimal $k$-shot setting, which achieved the highest exercise quality classification accuracy, was used in the following more advanced prompting techniques.

Reasoning-elicitation techniques are known to enhance deep-learning model performance \cite{xiong2023can} and can be categorized into two approaches: \textit{white-box} and \textit{black-box}. White-box methods access model weights or activations but are impractical for powerful yet closed-source LLMs. In contrast, \textit{black-box} approaches involve prompting an LLM to articulate its reasoning \cite{becker2024cycles} through techniques such as Chain-of-Thought (CoT) \cite{wei2022chain}, certainty \cite{xiong2023can}, and probability prompting \cite{gu2024probabilistic}, offering insights into the model's rationale behind classification. \textit{CoT prompting} guides the LLM to break down complex tasks into intermediate reasoning steps, enabling a more systematic and accurate problem-solving process while also offering insights into the model's thought process. \textit{Certainty elicitation} generates a numerical certainty score ranging from 0 to 1, alongside assessing exercise quality, indicating the model's confidence in the accuracy of its evaluation. \textit{Probability elicitation} outputs the probability that the exercise is correct, quantifying exercise quality and allowing flexibility in setting thresholds to balance recall and precision trade-offs \cite{gu2024probabilistic}.


\begin{table}[h]
\centering
\caption{
Prompting techniques for rehabilitation exercise quality assessment. Example of 2-shot prompting for the squat exercise, where four labeled examples (two correct, two incorrect) are provided to the LLM to evaluate a 5th test sample. Data sequences follow each prompt.
}
\label{tab:prompt}
\resizebox{\linewidth}{!}{%
\begin{tabular}{m{0.2\linewidth} m{0.8\linewidth}}
\toprule
\textbf{Technique} & \textbf{Prompt} \\ 
\midrule
Classification & \pbox{\linewidth}{Identify the label for the \textbf{5th} data sample below containing sequences of features extracted from Kinect data of the squat exercise. Ensure the output adheres to the output format: \textcolor{red}{"Label"}. \textcolor{red}{The label is either 'correct' or 'incorrect'}. \\<Data 1, Label 1: correct> ... <Data 4, Label 4: incorrect> <Data 5>} \\ 
\midrule
Chain-of-Thought & \pbox{\linewidth}{Identify the label for the \textbf{5th} data sample below containing sequences of features extracted from Kinect data of the squat exercise. Ensure the output adheres to the output format: \textcolor{red}{"Label, Reasoning"}. \textcolor{red}{Explain your reasoning step by step}. \\<Data 1, Label 1: correct> ... <Data 4, Label 4: incorrect> <Data 5>} \\
\midrule
Probability & \pbox{\linewidth}{Identify the label for the \textbf{5th} data sample below containing sequences of features extracted from Kinect data of the squat exercise. Ensure the output adheres to the output format: \textcolor{red}{"Probability"}. \textcolor{red}{Provide a probability score, where a higher score means a higher probability towards 'correct' and a lower score for 'incorrect'}. \\<Data 1, Label 1: correct> ... <Data 4, Label 4: incorrect> <Data 5>} \\  
\midrule
Certainty & \pbox{\linewidth}{Identify the label for the \textbf{5th} data sample below containing sequences of features extracted from Kinect data of the squat exercise. Ensure the output adheres to the output format: \textcolor{red}{"Label, Certainty"}. \textcolor{red}{Give a score between 0 and 1 for how certain you are in your classification}. \\<Data 1, Label 1: correct> ... <Data 4, Label 4: incorrect> <Data 5>} \\ 
\bottomrule
\end{tabular}%
}
\end{table}


\textit{Role-play prompting} \cite{shanahan2023role} was deployed to generate textual feedback by asking the LLM to complete the task while embodying a persona, such as a physiotherapist, to deliver succinct advice to improve exercise quality. Figure \ref{fig:textual_feedback} shows an example of role-play prompts and corresponding textual feedback.

\subsection{Evaluation Metrics}
\label{sec:metrics}
The performance of the proposed method for rehabilitation exercise quality assessment was quantitatively evaluated using the binary ground-truth labels provided in the two aforementioned datasets. The evaluation metrics used were accuracy, precision, recall, and F1 score for all prompting techniques, and also Area Under the Receiver Operating Characteristic Curve (AUC-ROC), Area Under the Precision-Recall Curve (AUC-PR) if probability was elicited. As no publicly available rehabilitation exercise dataset includes ground-truth textual feedback \cite{sardari2023artificial,ettefagh2024technological,brennan2019feedback}, the performance of the proposed method for feedback generation was assessed qualitatively.

\section{Experiments}
\label{sec:experiments}
\vspace{-2mm}
This section presents a quantitative evaluation of the proposed method for rehabilitation exercise quality assessment and a qualitative evaluation of textual feedback generation. The results are presented on two publicly available datasets REHAB24-6 and UI-PRMD. GPT-4o \cite{openai2024gpt4o} was used as the pre-trained LLM in all experiments.

\subsection{Datasets}
\label{sec:datasets}
UI-PRMD comprises 10 rehabilitation exercise types performed by 10 healthy subjects. Each subject completed 10 repetitions of each exercise, both correctly and incorrectly, on their dominant side. Body-joint data were captured using a Kinect sensor at 30 frames per second, with dataset samples corresponding to individual exercise repetitions. The 10 exercise types in UI-PRMD are outlined in Table \ref{tab:features} (a).

REHAB24-6 features data from 10 subjects performing 6 rehabilitation exercises, correctly and incorrectly. Body-joint data were collected using inertial wearable sensors. The dataset includes annotations on exercise correctness (correct vs. incorrect) and the start and end of exercise repetitions, enabling the creation of individual exercise repetition data samples. The 6 exercise types in REHAB24-6 are outlined in Table \ref{tab:features} (b).


\subsection{Experimental Results}
\label{sec:results}
\subsubsection{Few-shot Prompting}
\label{sec:shot-prompting}
\begin{figure}[h]
\centering \includegraphics[width=0.75\textwidth]{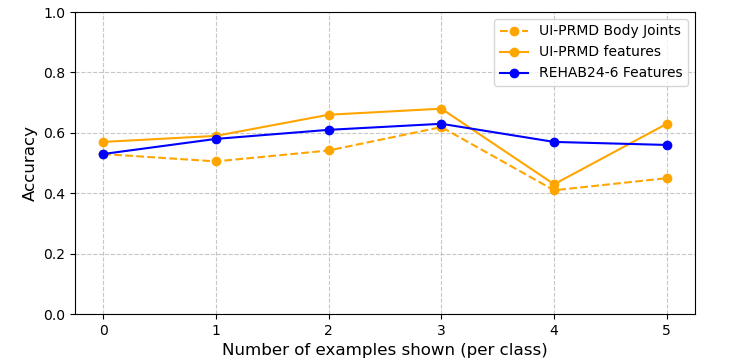}
\caption{Rehabilitation exercise quality classification accuracy varies with the number of labeled examples included in the prompts, evaluated on body joint data from UI-PRMD (orange dashed line), feature sequences extracted from UI-PRMD (orange solid line), and feature sequences extracted from REHAB24-6 (blue solid line).}
\label{fig:experiment1}
\end{figure}

Figure \ref{fig:experiment1} illustrates the accuracy of LLM-based exercise quality classification as the number of labeled examples increases, while Table \ref{tab:shot_results} presents the corresponding precision, recall, and F1 scores. It is evident that \textit{zero-shot prompting} yields the lowest accuracy, and incorporating more examples into the prompt generally enhances performance. However, beyond three-shot prompting, performance becomes inconsistent and may even decline. This observation aligns with previous findings \cite{brown2020language} that indicate that larger values of $k$ does not always lead to improved outcomes in LLMs. Consequently, few-shot prompting proves to be well-suited to the task of exercise assessment, particularly when datasets are limited in size. Unlike traditional machine learning approaches that require large training sets, few-shot prompting achieves comparable results with only a small number of examples.

Furthermore, Figure \ref{fig:experiment1} highlights the impact of input data type on classification accuracy. The results indicate that using exercise-specific features extracted from body joint data yields superior performance compared to using only raw body joint data. This improvement is due to the fact that extracted features are more interpretable and encapsulate domain-specific knowledge, as outlined in the feature extraction process detailed in Section \ref{sec:feature_extraction}. Additionally, extracted features have a significantly lower dimensionality than raw body joint data, effectively abstracting less relevant joints for a given movement.

\begin{table*}[h]
\centering
\caption{Rehabilitation exercise quality classification accuracy, precision, recall, and F1 score varies with the number of labeled examples included in the prompts provided to the LLM, evaluated on feature sequences extracted from (a) UI-PRMD and (b) REHAB24-6. Bolded values denote the best results.}
\label{tab:shot_results}
\begin{minipage}{0.48\linewidth}
    \centering
    \subcaption{UI-PRMD}
    \begin{tabular}{lcccc}
    \toprule
    \ & Accuracy & Precision & Recall & F1 \\ 
    \midrule
    0-shot        & 0.57 & 0.55 & 0.75 & 0.64 \\ 
    1-shot        & 0.59 & 0.57 & 0.72 & 0.63 \\ 
    2-shot        & 0.66 & 0.62 & \textbf{0.84} & 0.71 \\ 
    3-shot        & \textbf{0.68} & \textbf{0.74} & 0.79 & \textbf{0.76} \\ 
    4-shot        & 0.42 & 0.43 & 0.50 & 0.46 \\ 
    5-shot        & 0.63 & 0.64 & 0.60 & 0.62 \\ 
    \bottomrule
    \end{tabular}
\end{minipage}
\hfill
\begin{minipage}{0.48\linewidth}
    \centering
    \subcaption{REHAB24-6}
    \begin{tabular}{lcccc}
    \toprule
    \ & Accuracy & Precision & Recall & F1 \\ 
    \midrule
    0-shot        & 0.53 & 0.54 & 0.73 & 0.62 \\ 
    1-shot        & 0.58 & 0.58 & 0.77 & 0.66 \\ 
    2-shot        & 0.61 & 0.59 & 0.81 & 0.68 \\ 
    3-shot        & \textbf{0.63} & 0.60 & \textbf{0.85} & \textbf{0.70 }\\ 
    4-shot        & 0.57 & \textbf{0.68} & 0.72 & 0.65 \\ 
    5-shot        & 0.56 & 0.62 & 0.61 & 0.60 \\ 
    \bottomrule
    \end{tabular}
\end{minipage}
\end{table*}

\subsubsection{Reasoning Elicitation}
\label{sec:reasoning-elicitation}
Building on the best-performing setting from the few-shot prompting experiments, three-shot prompting with feature sequences was selected for reasoning elicitation and subsequent experiments. CoT, certainty, probability, and a combination of CoT-with-certainty were evaluated on the two datasets. As shown in Table \ref{tab:reasoning_results}, reasoning-elicitation methods generally outperformed baseline prompting. Among these methods, CoT and certainty prompting emerged as the most effective strategies. Notably, while accuracy scores remained similar for both approaches, slight differences were observed in precision and recall.

\begin{table*}[h]
\centering
\caption{Performance of the LLM with different prompting techniques on (a) UI-PRMD and (b) REHAB24-6. The results of LSTM and ST-GCN are presented for (b) REHAB24-6. Among the prompting techniques, only probability elicitation can generate class probability estimates, allowing for the computation of AUC-ROC and AUC-PR.}
\label{tab:reasoning_results}
\begin{tabular}{llcccccc}
\toprule
\multicolumn{7}{l}{(a) UI-PRMD} \\ 
\midrule
Setting & Accuracy & Precision & Recall & F1 & AUC-ROC & AUC-PR \\ 
\midrule
3-shot        & 0.68 & 0.74 & 0.79 & 0.76 & - & - \\ 
Chain-of-Thought           & 0.72 & 0.75 & 0.67 & 0.71 & - & - \\ 
Certainty     & 0.76 & 0.72 & 0.87 & 0.79 & - & - \\ 
Probability   & 0.68 & 0.65 & 0.79 & 0.71 & 0.70 & 0.68 \\ 
Chain-of-Thought + Certainty   & 0.64 & 0.59 & \textbf{0.90} & 0.72 & - & - \\
LSTM \cite{abedi2023cross}   & 0.87    & 0.97    & 0.88    & 0.92    & 0.97  & 0.96  \\ 
ST-GCN \cite{zheng2023skeleton} & \textbf{0.94}    & \textbf{0.98}    & \textbf{0.90}    & \textbf{0.96}    & \textbf{0.98}  & \textbf{0.98}  \\ 
\midrule
\multicolumn{7}{l}{(b) REHAB24-6} \\ 
\midrule
Setting & Accuracy & Precision & Recall & F1 & AUC-ROC & AUC-PR \\  
\midrule
3-shot        & 0.63 & 0.60 & 0.85 & 0.70 & - & - \\
Chain-of-Thought           & \textbf{0.70} & \textbf{0.71} & 0.67 & 0.69 & - & - \\ 
Certainty     & \textbf{0.70} & 0.67 & \textbf{0.80} & \textbf{0.73} & - & - \\ 
Probability   & 0.67 & 0.63 & \textbf{0.80} & 0.71 & 0.72 & 0.68 \\ 
Chain-of-Thought + Certainty   & 0.67 & 0.63 & \textbf{0.80} & 0.71 & - & - \\
LSTM \cite{abedi2023cross}   & 0.60    & 0.63    & 0.60    & 0.61    & 0.64  & 0.70  \\ 
ST-GCN \cite{zheng2023skeleton} & 0.63    & 0.61    & 0.83    & 0.70    & 0.69  & 0.64  \\ 
\bottomrule
\end{tabular}
\label{tab:performance}
\end{table*}

CoT has higher precision but lower recall than certainty, suggesting CoT is more selective in classifying movements as correct. On the other hand, certainty has higher recall but lower precision than CoT, indicating more movements are classified as correct, which includes more false positives. This trade-off should be considered in the real-world implementation of this system, should healthcare professionals want to adjust sensitivity levels.

Despite their individual strengths, combining CoT with certainty did not yield improved results. Certainty-based prompting has similar effects to CoT reasoning in LLM classification tasks, as it implicitly prompts the model to justify its predictions. This suggests that certainty prompting slightly outperforms CoT by reducing CoT-induced "hallucinations" \cite{becker2024cycles}. In CoT experiments, erroneous reasoning was found to lead to an accumulation of mistakes.  Additionally, zero-shot CoT sometimes caused the LLM to create its own classification thresholds, which could either exaggerate errors or misalign with rehabilitation clinicians' expectations. For instance, the LLM outputted: "the shoulder abduction angle reaches a maximum of 160°, which is significantly higher than the expected 150°." This not only exaggerated the mistake but also imposed a threshold of 150°, whereas UI-PRMD physiotherapists may consider shoulder abductions non-optimal when the patient exhibits less than 160° of abduction. Consequently, the LLM would classify subsequent samples within the same conversation based on this initially self-assigned hard threshold. This trend was most pronounced when CoT reasoning was explicitly extracted. This highlights a key motivation for using LLMs rather than fixed threshold-based algorithms, as rehabilitation assessment requires adaptive interpretation of patient movement patterns, which may vary significantly across patients. Additionally, the reasoning-elicitation experiments reveal that likely explanations are not always correct, as most of the certainty and probability scores generated by the LLM ranged between 0.8 and 1, regardless of the actual accuracy of the assessment. This indicates that the broader issue of overconfidence in LLMs persists in the context of exercise quality assessment \cite{xiong2023can,becker2024cycles}.

To compare the performance of the LLM with previous deep learning techniques \cite{abedi2023cross,zheng2023skeleton}, the last two rows of Table \ref{tab:reasoning_results} (a) and (b) present the results of LSTM and ST-GCN on UI-PRMD and REHAB24-6. The LSTM model features a two-layer architecture, with each layer containing 64 hidden units, followed by a fully connected layer of size $64 \times 1$, where the single output dimension corresponds to the binary classification task. The ST-GCN model comprises three ST-GCN layers, followed by an average pooling layer, as designed in \cite{zheng2023skeleton}. The output from the pooling layer is further processed by a convolutional layer that maps it to a single output dimension for classification. As shown in Table \ref{tab:reasoning_results} (a) and (b), while the LLM, with any prompting technique, outperformed both LSTM and ST-GCN for REHAB24-6, its performance was inferior to that of LSTM and ST-GCN for UI-PRMD. The higher inter-class separability of data samples in UI-PRMD compared to REHAB24-6 reduces classification complexity, making it more suitable for traditional deep-learning models. Although the LLM performs worse than LSTM and ST-GCN on UI-PRMD, it is still advantageous by providing feedback and reasoning on correct and incorrect classifications, enhancing interpretability beyond conventional approaches.

\subsection{Exercise-specific Results Across Datasets}
\label{sec:classification_evaluation}

Building on the optimal prompt settings from previous experiments, three-shot prompting with extracted features and certainty elicitation was tested across common movements in REHAB24-6 and UI-PRMD.

\begin{table*}[h]
    \centering
    \caption{Performance comparison of GPT-4o on rehabilitation exercise quality assessment for the three common exercises in REHAB24-6 and UI-PRMD: shoulder abduction (ex1, m07), leg lunge (ex5, m03), and squat (ex6, m01), refer to Table \ref{tab:features}.}
    \label{tab:classification_results}
    \begin{tabular}{lccc|ccc}
        \hline
        & \multicolumn{3}{c|}{REHAB24-6} & \multicolumn{3}{c}{UI-PRMD} \\
        \hline
        Exercise & ex1 & ex5 & ex6 & m07 & m03 & m01 \\
        \midrule
        Accuracy & 0.67 & 0.74 & \textbf{0.75} & \textbf{0.76} & 0.67 & \textbf{0.76}\\
        Precision & 0.71 & 0.69 & \textbf{0.78} & \textbf{0.76} & 0.64 & 0.69 \\
        Recall & 0.75 & \textbf{0.90} & 0.70 & 0.84 & 0.76 & \textbf{0.95} \\
        F1 & 0.73 & \textbf{0.78} & 0.74 & \textbf{0.80} & 0.70 & \textbf{0.80} \\
        \bottomrule
    \end{tabular}
\end{table*}

Differences in performance across different movements were observed in Table \ref{tab:classification_results}, likely due to inherent variations in extracted features and their effectiveness in capturing movement errors. For both datasets, the squat exercise (ex6, m01) was particularly well-suited for LLM evaluation. The inherent symmetry of this movement likely creates a better distinction between correct and incorrect executions. In contrast, exercises such as leg lunges and shoulder abductions primarily involve the movement of a single body part, potentially leading to oversight of other contributing factors to incorrect movements. For instance, if a patient performs a leg lunge but exhibits subtle instability in their upper body, which is not explicitly included in the analysis, the LLM may classify the movement as correct, whereas a rehabilitation expert would identify the movement as incorrect.


\subsubsection{Feedback Evaluation}
\label{sec:feedback_evaluation}
\begin{figure}[h]
\centering
\includegraphics[width=\textwidth]{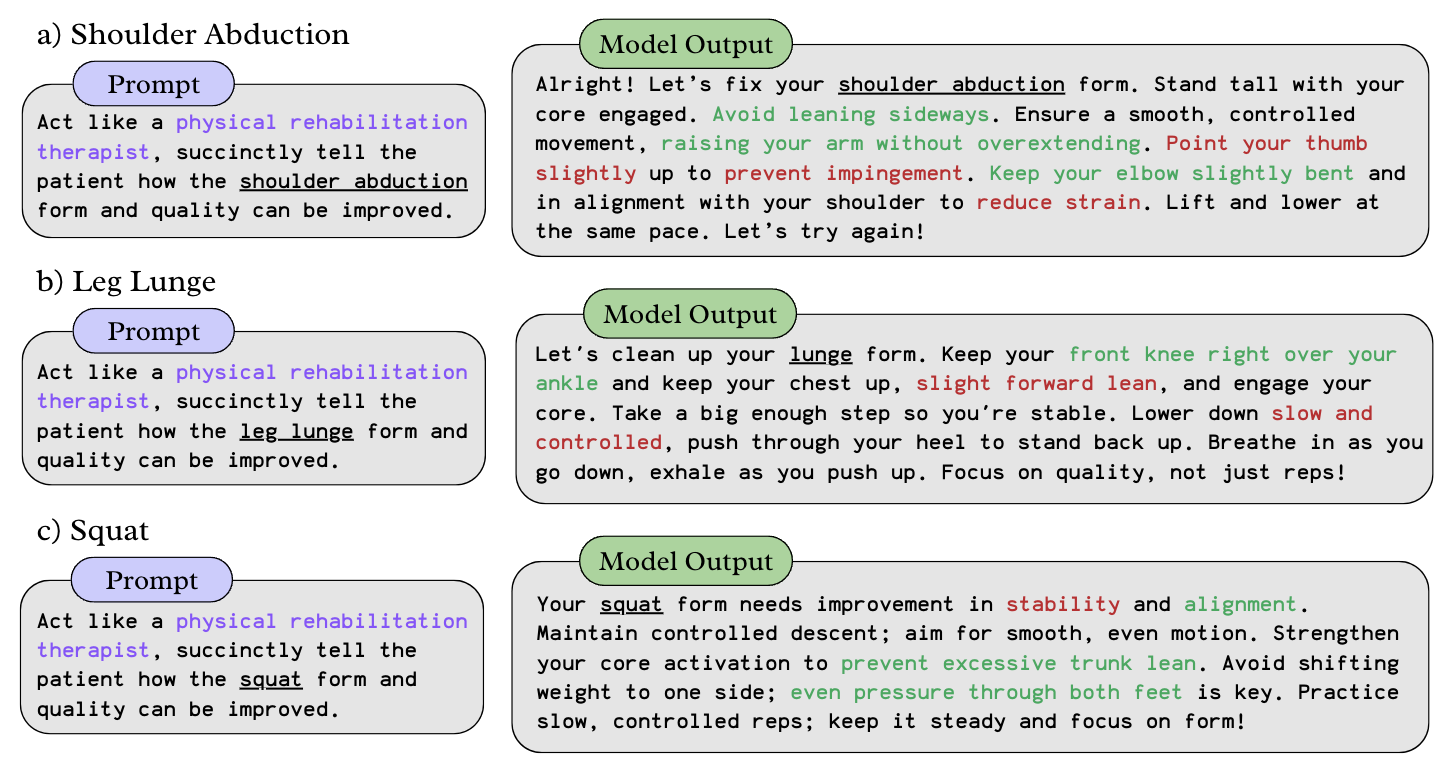}
\caption{Role-playing prompts for feedback generation following classification and reasoning, applied to (a) shoulder abduction, (b) leg lunge, and (c) squat exercise. The corresponding classification results are presented in Table \ref{tab:classification_results}. The LLM generates textual feedback in the specified role’s style, incorporating suggestions derived from data trends (highlighted in green) and insights from its prior knowledge (highlighted in red).}
\label{fig:textual_feedback}
\end{figure}

After obtaining the exercise quality prediction from the LLM, a role-play prompt was employed to generate concise textual feedback. Providing textual feedback after the model has already assessed exercise quality offers additional insight into its reasoning process. As illustrated in Figure \ref{fig:textual_feedback}, prompting the LLM to explain how the patient can correct their movement by utilizing the role-playing technique within our two-step framework. This approach has been shown to enhance trust in initial responses and improve certainty estimations \cite{li2024think} by granting the LLM a second opportunity for reasoning and justification.  Moreover, the LLM-generated feedback not only directly addresses the extracted features but also incorporates prior knowledge and general movement expectations. These include considerations such as stability, controlled movement speed, reducing strain, and breathing techniques, further enriching the quality of the feedback.

\section{Conclusion and Future Works}
\label{sec:conclusion}
This paper proposed a novel method utilizing LLMs for rehabilitation exercise quality assessment and feedback generation. By integrating advanced prompting techniques with exercise-specific features extracted from body joint sequences, the method enabled LLMs to analyze features, assess exercise quality, and generate meaningful feedback. The proposed approach achieved high accuracy in exercise assessments with few-shot prompting and produced valuable feedback. We also observed that the LLM remained overconfident of their outcomes, irrespective of whether the decision was correct or not. Our results demonstrate the potential of LLMs to support accurate, explainable, and adaptable AI-driven virtual rehabilitation systems. Despite promising results, the proposed approach has certain limitations. One key limitation is the lack of quantitative evaluation for feedback generation due to the absence of publicly available rehabilitation exercise datasets with ground-truth textual feedback. Additionally, the method relies on pre-trained LLMs, with data analysis performed based on their knowledge base. While prompting techniques such as few-shot prompting improve the LLMs' ability to evaluate exercises, fine-tuning LLMs specifically for this task could further enhance performance. Furthermore, the inability to set a random seed in the GPT-4o LLM affects result reproducibility. Since the GPT-4o LLM does not allow for controlled randomness, the generated feedback may vary across different runs, making it challenging to ensure consistent evaluation and comparison across different experiments. Future work will focus on collecting and annotating rehabilitation exercise datasets with ground-truth feedback and leveraging this data to fine-tune pre-trained LLMs, enhancing their performance and reliability in exercise quality assessment and feedback generation.

\vspace{8mm}
\noindent
\textbf{Funding--}This research was funded by the New Frontiers in Research Fund, Canada, and the TRANSFORM HF Undergraduate Summer Research Program, Canada.

\begingroup
\setstretch{1.0} 
\bibliographystyle{IEEEtran}
\bibliography{references}

\begin{thebibliography}{10}
\providecommand{\url}[1]{#1}
\csname url@samestyle\endcsname
\providecommand{\newblock}{\relax}
\providecommand{\bibinfo}[2]{#2}
\providecommand{\BIBentrySTDinterwordspacing}{\spaceskip=0pt\relax}
\providecommand{\BIBentryALTinterwordstretchfactor}{4}
\providecommand{\BIBentryALTinterwordspacing}{\spaceskip=\fontdimen2\font plus
\BIBentryALTinterwordstretchfactor\fontdimen3\font minus \fontdimen4\font\relax}
\providecommand{\BIBforeignlanguage}[2]{{%
\expandafter\ifx\csname l@#1\endcsname\relax
\typeout{** WARNING: IEEEtran.bst: No hyphenation pattern has been}%
\typeout{** loaded for the language `#1'. Using the pattern for}%
\typeout{** the default language instead.}%
\else
\language=\csname l@#1\endcsname
\fi
#2}}
\providecommand{\BIBdecl}{\relax}
\BIBdecl

\bibitem{WHORehabilitation}
{World Health Organization}, ``{Rehabilitation},'' \url{https://www.who.int/news-room/fact-sheets/detail/rehabilitation}, {2023}, {Accessed: January 30, 2023}.

\bibitem{dibben2023exercise}
G.~O. Dibben, J.~Faulkner, N.~Oldridge, K.~Rees, D.~R. Thompson, A.-D. Zwisler, and R.~S. Taylor, ``Exercise-based cardiac rehabilitation for coronary heart disease: a meta-analysis,'' \emph{European heart journal}, vol.~44, no.~6, pp. 452--469, 2023.

\bibitem{shirozhan2022barriers}
S.~Shirozhan, N.~Arsalani, S.~S.~B. Maddah, and F.~Mohammadi-Shahboulaghi, ``Barriers and facilitators of rehabilitation nursing care for patients with disability in the rehabilitation hospital: A qualitative study,'' \emph{Frontiers in Public Health}, vol.~10, 2022.

\bibitem{ferreira2023usage}
R.~Ferreira, R.~Santos, and A.~Sousa, ``Usage of auxiliary systems and artificial intelligence in home-based rehabilitation: A review,'' \emph{Exploring the Convergence of Computer and Medical Science Through Cloud Healthcare}, pp. 163--196, 2023.

\bibitem{seron2021effectiveness}
P.~Seron, M.-J. Oliveros, R.~Gutierrez-Arias, R.~Fuentes-Aspe, R.~C. Torres-Castro, C.~Merino-Osorio, P.~Nahuelhual, J.~Inostroza, Y.~Jalil, R.~Solano \emph{et~al.}, ``Effectiveness of telerehabilitation in physical therapy: a rapid overview,'' \emph{Physical therapy}, vol. 101, no.~6, p. pzab053, 2021.

\bibitem{boukhennoufa2022wearable}
I.~Boukhennoufa, X.~Zhai, V.~Utti, J.~Jackson, and K.~D. McDonald-Maier, ``Wearable sensors and machine learning in post-stroke rehabilitation assessment: A systematic review,'' \emph{Biomedical Signal Processing and Control}, vol.~71, p. 103197, 2022.

\bibitem{abedi2024artificial}
A.~Abedi, T.~J. Colella, M.~Pakosh, and S.~S. Khan, ``Artificial intelligence-driven virtual rehabilitation for people living in the community: A scoping review,'' \emph{NPJ Digital Medicine}, vol.~7, no.~1, p.~25, 2024.

\bibitem{sardari2023artificial}
S.~Sardari, S.~Sharifzadeh, A.~Daneshkhah, B.~Nakisa, S.~W. Loke, V.~Palade, and M.~J. Duncan, ``Artificial intelligence for skeleton-based physical rehabilitation action evaluation: A systematic review,'' \emph{Computers in Biology and Medicine}, p. 106835, 2023.

\bibitem{ettefagh2024technological}
A.~Ettefagh and A.~Roshan~Fekr, ``Technological advances in lower-limb tele-rehabilitation: A review of literature,'' \emph{Journal of Rehabilitation and Assistive Technologies Engineering}, vol.~11, p. 20556683241259256, 2024.

\bibitem{brennan2019feedback}
L.~Brennan, E.~Dorronzoro~Zubiete, and B.~Caulfield, ``Feedback design in targeted exercise digital biofeedback systems for home rehabilitation: A scoping review,'' \emph{Sensors}, vol.~20, no.~1, p. 181, 2019.

\bibitem{pavllo20193d}
D.~Pavllo, C.~Feichtenhofer, D.~Grangier, and M.~Auli, ``3d human pose estimation in video with temporal convolutions and semi-supervised training,'' in \emph{Proceedings of the IEEE/CVF conference on computer vision and pattern recognition}, 2019, pp. 7753--7762.

\bibitem{lugaresi2019mediapipe}
C.~Lugaresi, J.~Tang, H.~Nash, C.~McClanahan, E.~Uboweja, M.~Hays, F.~Zhang, C.-L. Chang, M.~G. Yong, J.~Lee \emph{et~al.}, ``Mediapipe: A framework for building perception pipelines,'' \emph{arXiv preprint arXiv:1906.08172}, 2019.

\bibitem{abedi2023cross}
A.~Abedi, M.~Malmirian, and S.~S. Khan, ``Cross-modal video to body-joints augmentation for rehabilitation exercise quality assessment,'' \emph{arXiv preprint arXiv:2306.09546}, 2023.

\bibitem{yan2018spatial}
S.~Yan, Y.~Xiong, and D.~Lin, ``Spatial temporal graph convolutional networks for skeleton-based action recognition,'' in \emph{Proceedings of the AAAI conference on artificial intelligence}, vol.~32, no.~1, 2018.

\bibitem{kimore}
M.~Capecci, M.~Ceravolo, F.~Ferracuti, S.~Iarlori, A.~Monteriu, L.~Romeo, and F.~Verdini, ``The kimore dataset: Kinematic assessment of movement and clinical scores for remote monitoring of physical rehabilitation,'' \emph{IEEE Transactions on Neural Systems and Rehabilitation Engineering}, vol.~27, no.~7, pp. 1436--1448, 2019, epub 2019 Jun 14.

\bibitem{schuber2024relevance}
A.~A. Schuber and A.~Schaller, ``Relevance of therapist feedback in the context of group-based exercise programs in medical rehabilitation--results from a qualitative study with patients and exercise therapists,'' \emph{European Journal of Physiotherapy}, pp. 1--9, 2024.

\bibitem{popovic2014feedback}
M.~D. Popovi{\'c}, M.~D. Kosti{\'c}, S.~Z. Rodi{\'c}, and L.~M. Konstantinovi{\'c}, ``Feedback-mediated upper extremities exercise: Increasing patient motivation in poststroke rehabilitation,'' \emph{BioMed research international}, vol. 2014, no.~1, p. 520374, 2014.

\bibitem{wang2024ubiphysio}
C.~Wang, Y.~Feng, L.~Zhong, S.~Zhu, C.~Zhang, S.~Zheng, C.~Liang, Y.~Wang, C.~He, C.~Yu \emph{et~al.}, ``Ubiphysio: Support daily functioning, fitness, and rehabilitation with action understanding and feedback in natural language,'' \emph{Proceedings of the ACM on Interactive, Mobile, Wearable and Ubiquitous Technologies}, vol.~8, no.~1, pp. 1--27, 2024.

\bibitem{prmd}
\BIBentryALTinterwordspacing
A.~Vakanski, H.-p. Jun, D.~Paul, and R.~Baker, ``A data set of human body movements for physical rehabilitation exercises,'' \emph{Data}, vol.~3, no.~1, 2018. [Online]. Available: \url{https://www.mdpi.com/2306-5729/3/1/2}
\BIBentrySTDinterwordspacing

\bibitem{vcernek2024rehab24}
A.~{\v{C}}ernek, J.~Sedmidubsky, and P.~Budikova, ``Rehab24-6: Physical therapy dataset for analyzing pose estimation methods,'' in \emph{International Conference on Similarity Search and Applications}.\hskip 1em plus 0.5em minus 0.4em\relax Springer, 2024, pp. 18--33.

\bibitem{liao2020deep}
Y.~Liao, A.~Vakanski, and M.~Xian, ``A deep learning framework for assessing physical rehabilitation exercises,'' \emph{IEEE Transactions on Neural Systems and Rehabilitation Engineering}, vol.~28, no.~2, pp. 468--477, 2020.

\bibitem{guo2021exercise}
Q.~Guo and S.~S. Khan, ``Exercise-specific feature extraction approach for assessing physical rehabilitation,'' in \emph{4th IJCAI Workshop on AI for Aging, Rehabilitation and Intelligent Assisted Living. IJCAI}, 2021.

\bibitem{karagoz2023supervised}
B.~Karagoz, A.~Ashraf, and S.~Khan, ``Supervised sequential contrastive regression: Improving performance on imbalanced rehabilitation exercises datasets,'' \emph{preprint}, 12 2023.

\bibitem{deb2022graph}
S.~Deb, M.~F. Islam, S.~Rahman, and S.~Rahman, ``Graph convolutional networks for assessment of physical rehabilitation exercises,'' \emph{IEEE Transactions on Neural Systems and Rehabilitation Engineering}, vol.~30, pp. 410--419, 2022.

\bibitem{zheng2023skeleton}
K.~Zheng, J.~Wu, J.~Zhang, and C.~Guo, ``A skeleton-based rehabilitation exercise assessment system with rotation invariance,'' \emph{IEEE Transactions on Neural Systems and Rehabilitation Engineering}, 2023.

\bibitem{reby2023graph}
K.~R{\'e}by, I.~Dulau, G.~Dubrasquet, and M.~B. Aimar, ``Graph transformer for physical rehabilitation evaluation,'' in \emph{2023 IEEE 17th International Conference on Automatic Face and Gesture Recognition (FG)}.\hskip 1em plus 0.5em minus 0.4em\relax IEEE, 2023, pp. 1--8.

\bibitem{karlov2024rehabilitation}
M.~Karlov, A.~Abedi, and S.~S. Khan, ``Rehabilitation exercise quality assessment through supervised contrastive learning with hard and soft negatives,'' \emph{Medical \& Biological Engineering \& Computing}, pp. 1--14, 2024.

\bibitem{bruce2024egcn++}
X.~Bruce, Y.~Liu, K.~C. Chan, and C.~W. Chen, ``Egcn++: A new fusion strategy for ensemble learning in skeleton-based rehabilitation exercise assessment,'' \emph{IEEE Transactions on Pattern Analysis and Machine Intelligence}, 2024.

\bibitem{parker2011review}
J.~Parker, G.~Mountain, and J.~Hammerton, ``A review of the evidence underpinning the use of visual and auditory feedback for computer technology in post-stroke upper-limb rehabilitation,'' \emph{Disability and rehabilitation: Assistive technology}, vol.~6, no.~6, pp. 465--472, 2011.

\bibitem{li2024finerehab}
J.~Li, J.~Xue, R.~Cao, X.~Du, S.~Mo, K.~Ran, and Z.~Zhang, ``Finerehab: A multi-modality and multi-task dataset for rehabilitation analysis,'' in \emph{Proceedings of the IEEE/CVF Conference on Computer Vision and Pattern Recognition}, 2024, pp. 3184--3193.

\bibitem{irds}
\BIBentryALTinterwordspacing
A.~Miron, N.~Sadawi, W.~Ismail, H.~Hussain, and C.~Grosan, ``Intellirehabds (irds)—a dataset of physical rehabilitation movements,'' \emph{Data}, vol.~6, no.~5, 2021. [Online]. Available: \url{https://www.mdpi.com/2306-5729/6/5/46}
\BIBentrySTDinterwordspacing

\bibitem{zaghir2024prompt}
J.~Zaghir, M.~Naguib, M.~Bjelogrlic, A.~N{\'e}v{\'e}ol, X.~Tannier, and C.~Lovis, ``Prompt engineering paradigms for medical applications: Scoping review,'' \emph{Journal of Medical Internet Research}, vol.~26, p. e60501, 2024.

\bibitem{brown2020language}
T.~Brown, B.~Mann, N.~Ryder, M.~Subbiah, J.~D. Kaplan, P.~Dhariwal, A.~Neelakantan, P.~Shyam, G.~Sastry, A.~Askell \emph{et~al.}, ``Language models are few-shot learners,'' \emph{Advances in neural information processing systems}, vol.~33, pp. 1877--1901, 2020.

\bibitem{touvron2023llama}
H.~Touvron, T.~Lavril, G.~Izacard, X.~Martinet, M.-A. Lachaux, T.~Lacroix, B.~Rozi{\`e}re, N.~Goyal, E.~Hambro, F.~Azhar \emph{et~al.}, ``Llama: Open and efficient foundation language models,'' \emph{arXiv preprint arXiv:2302.13971}, 2023.

\bibitem{xiong2023can}
M.~Xiong, Z.~Hu, X.~Lu, Y.~Li, J.~Fu, J.~He, and B.~Hooi, ``Can llms express their uncertainty? an empirical evaluation of confidence elicitation in llms,'' \emph{arXiv preprint arXiv:2306.13063}, 2023.

\bibitem{becker2024cycles}
E.~Becker and S.~Soatto, ``Cycles of thought: Measuring llm confidence through stable explanations,'' \emph{arXiv preprint arXiv:2406.03441}, 2024.

\bibitem{wei2022chain}
J.~Wei, X.~Wang, D.~Schuurmans, M.~Bosma, F.~Xia, E.~Chi, Q.~V. Le, D.~Zhou \emph{et~al.}, ``Chain-of-thought prompting elicits reasoning in large language models,'' \emph{Advances in neural information processing systems}, vol.~35, pp. 24\,824--24\,837, 2022.

\bibitem{gu2024probabilistic}
B.~Gu, R.~J. Desai, K.~J. Lin, and J.~Yang, ``Probabilistic medical predictions of large language models,'' \emph{npj Digital Medicine}, vol.~7, no.~1, p. 367, 2024.

\bibitem{shanahan2023role}
M.~Shanahan, K.~McDonell, and L.~Reynolds, ``Role play with large language models,'' \emph{Nature}, vol. 623, no. 7987, pp. 493--498, 2023.

\bibitem{openai2024gpt4o}
\BIBentryALTinterwordspacing
OpenAI, ``Gpt-4o announcement,'' 2024, accessed: 2025-02-17. [Online]. Available: \url{https://openai.com/index/hello-gpt-4o}
\BIBentrySTDinterwordspacing

\bibitem{li2024think}
M.~Li, W.~Wang, F.~Feng, F.~Zhu, Q.~Wang, and T.-S. Chua, ``Think twice before assure: Confidence estimation for large language models through reflection on multiple answers,'' \emph{arXiv preprint arXiv:2403.09972}, 2024.

\end{thebibliography}
\endgroup

\end{document}